%% file: emnlp2021.tex
\pdfoutput=1

\documentclass[11pt]{article}

\usepackage[]{emnlp2021}

\usepackage{times}
\usepackage{latexsym}

\usepackage[T1]{fontenc}

\usepackage[utf8]{inputenc}

\usepackage{microtype}

\usepackage{amsmath}
\usepackage{cleveref}
\usepackage{appendix}
\usepackage{booktabs}
\usepackage{tablefootnote}
\usepackage{mdframed}
\usepackage{textcomp}
\usepackage{graphicx}
\usepackage{arydshln}
\usepackage{subcaption}
\usepackage{xspace}

\newcommand{\abr}[1]{\textsc{#1}}
\newcommand{\respace}{}

\newcommand\original{\textit{Original}\xspace}
\newcommand\substitute{\textit{Substitute}\xspace}
\newcommand\other{\textit{Other}\xspace}

\newif\ifcomments
\commentstrue 
\ifcomments
    \providecommand{\anthony}[2][]{{\protect\color{blue}{[anthony:\textbf{#1} #2]}}}
    \providecommand{\erik}[2][]{{\protect\color{red}{[erik:\textbf{#1} #2]}}}
    \providecommand{\ej}[2][]{{\protect\color{red}{[erik:\textbf{#1} #2]}}}
    \providecommand{\sameer}[2][]{{\protect\color{magenta}{[sameer:\textbf{#1} #2]}}}
    \providecommand{\kartik}[2][]{{\protect\color{red}{[kartik:\textbf{#1} #2]}}}
\else
    \providecommand{\anthony}[2][]{}
    \providecommand{\erik}[2][]{}
    \providecommand{\ej}[2][]{}
    \providecommand{\sameer}[2]{}
    \providecommand{\kartik}[2][]{}
\fi

\newcommand{\nerType}[1]{\texttt{#1}\xspace}
\newcommand{\PER}{\nerType{PER}}
\newcommand{\DAT}{\nerType{DAT}}
\newcommand{\NUM}{\nerType{NUM}}
\newcommand{\ORG}{\nerType{ORG}}
\newcommand{\LOC}{\nerType{LOC}}
\newcommand*\samethanks[1][\value{footnote}]{\footnotemark[#1]}
\newcommand{\authorspace}{\hspace{0.3cm}}

\title{Entity-Based Knowledge Conflicts in Question Answering}

\author{
  \bf Shayne Longpre\thanks{~~Equal Contribution.}$\:\,^{\spadesuit}$\authorspace{}
  \bf Kartik Perisetla\samethanks$\:\,^{\spadesuit}$ \authorspace{}
  \bf Anthony Chen\samethanks$\:\,^{\heartsuit}$ 
  \\
  \bf Nikhil Ramesh$^{\spadesuit}$ \authorspace
  \bf Chris DuBois$^{\spadesuit}$ \authorspace
  \bf Sameer Singh$^{\heartsuit}$  \vspace{2mm}
  \\
  $^{\spadesuit}$Apple \authorspace{}
  $^{\heartsuit}$University of California, Irvine\\
  \href{mailto:slongpre@mit.edu}{\tt slongpre@mit.edu} 
  \\
  \{\href{mailto:kperisetla@apple.com}{\tt kperisetla},
    \href{mailto:nikhilr@apple.com}{\tt nikhilr},
    \href{mailto:cdubois@apple.com}{\tt cdubois}\}
    \href{mailto:kperisetla@apple.com}{\tt @apple.com}
  \\
  \{\href{mailto:anthony.chen@uci.edu}{\tt anthony.chen},
    \href{mailto:sameer@uci.edu}{\tt sameer}\}
    \href{mailto:anthony.chen@uci.edu}{\tt @uci.edu}
}

\begin{document}
\maketitle

\input{sections/0-abstract}

\input{sections/1-introduction-v2}
\input{sections/2-sub-framework}

\input{sections/3-methods}
\input{sections/4-experiments}
\input{sections/6-relatedwork}
\input{sections/7-conclusion}

\respace
\respace
\section*{Acknowledgements}
\respace
\respace
We thank Ni Lao, Yu Wang, Russ Webb, Adam Fisch, Matt Gardner, Dheeru Dua, Sanjay Subramanian, and the anonymous reviewers for their valuable feedback.
This work is funded in part by the DARPA MCS program under Contract No. N660011924033 with the United States Office Of Naval Research and in part by NSF award \#IIS-1817183.

\bibliography{anthology,custom}
\bibliographystyle{acl_natbib}

\end{document}

%% file: sections/0-abstract.tex
\begin{abstract}
Knowledge-dependent tasks typically use two sources of knowledge: \textit{parametric}, learned at training time, and \textit{contextual}, given as a passage at inference time.
To understand how models use these sources together, we formalize the problem of knowledge conflicts, where the contextual information contradicts the learned information.
Analyzing the behaviour of popular models, we measure their over-reliance on memorized information (the cause of hallucinations), and uncover important factors that exacerbate this behaviour.
Lastly, we propose a simple method to mitigate over-reliance on parametric knowledge which minimizes hallucination and improves out-of-distribution generalization by $4\% - 7\%$.
Our findings demonstrate the importance for practitioners to evaluate model tendency to hallucinate rather than read, and show that our mitigation strategy encourages generalization to evolving information (\textit{i.e.,} time-dependent queries).
To encourage these practices, we have released our framework for generating knowledge conflicts.\footnote{Framework is provided at \url{https://github.com/apple/ml-knowledge-conflicts}.}

\end{abstract}

%% file: sections/1-introduction-v2.tex
\section{Introduction}
\respace
Knowledge-dependent tasks, such as open-retrieval question answering (QA), require expansive ``world knowledge'', common sense, and reasoning abilities.
State-of-the-art approaches typically follow a retrieve-and-read setup \citep{chen2017reading}, where the retriever sources relevant documents, and the reader produces an answer from these.
In this sense, there are two sources of knowledge contributing to model inference with an ambiguous and opaque division of labour.
The first is the implicit parametric knowledge (\textit{i.e.}, their learned weights) instilled by pre-training and fine-tuning \citep{petroni2019language}. 
The second is contextual knowledge, usually sourced as passages of text from the retriever \citep{fisch2019mrqa}.

\begin{figure}[t!]
    \begin{mdframed}[linecolor=red!75!black,linewidth=1pt]
        \small \textbf{Question:} Who did US fight in world war 1?\\
        \textbf{Original Context:} The United States declared war on \textbf{\textcolor{red!75!black}{Germany}} on April 6, 1917, over 2 years after World War I started \ldots \\
        \textbf{Original Answer:} \textbf{\textcolor{red!75!black}{Germany}}
        
    \end{mdframed}
    \small \emph{\textbf{Model Prediction:} \textbf{\textcolor{red!75!black}{Germany}}} \\
    
    \begin{mdframed}[linecolor=green!50!black,linewidth=1pt]
        \small \textbf{Question:} Who did US fight in world war 1?\\
        \textbf{Substitute Context:} The United States declared war on \textbf{\textcolor{green!50!black}{Taiwan}} on April 6, 1917, over 2 years after World War I started \ldots \\
        \textbf{Substitute Answer:} \textbf{\textcolor{green!50!black}{Taiwan}}
        
    \end{mdframed}
    \small \emph{\textbf{Model Prediction:} \textbf{\textcolor{red!75!black}{Germany}}}
    
    \caption{\textbf{Knowledge Substitution:} A \textcolor{green!50!black}{\textbf{substitute example}} is derived from the \textcolor{red!75!black}{\textbf{original example}} by replacing the original answer, \textcolor{red!75!black}{\textbf{Germany}}, with a similar type of answer, \textit{i.e.} \textcolor{green!50!black}{\textbf{Taiwan}}.
    An example of a \textbf{knowledge conflict} occurs when a model is trained (or pre-trained) on the \textcolor{red!75!black}{\textbf{original example}} and evaluated on the \textcolor{green!50!black}{\textbf{substitute example}}.}
    \label{fig:introduction:example}
    \vspace*{-.6cm}
\end{figure}  
\respace
\respace

As a testament to their memorization abilities, large language models can produce competitive results relying only on their own parametric knowledge, without access to relevant documents \citep{Brown2020LanguageMA, roberts2020much}.
However, this memorization behaviour has manifested in a penchant to \textit{hallucinate}, or parrot answers memorized during training, completely ignoring relevant documents when provided \citep{Krishna2021HurdlesTP, 10.1145/3442188.3445922}.
This memorization behaviour violates the expectation that the reader produce answers consistent with the retrieved information, diminishing interpretability of the system.
More problematically, this behaviour inhibits the model's ability to generalize to evolving knowledge and time-dependent answers, not found in training \citep{guu2020realm, Schuster2021GetYV}.

Our objective is to understand how systems employ parametric and contextual knowledge together by studying knowledge conflicts: situations where the contextual knowledge contradicts with knowledge learned during pre-training or fine-tuning.
Because the space of knowledge conflicts is broad, we restrict ourselves to the space of \textit{entity-based} conflicts -- restricted to named entity substitutions.
We create an automated framework that identifies QA instances with named entity answers, then substitutes mentions of the entity in the gold document with an alternate entity, thus changing the answer (Fig. \ref{fig:introduction:example}).
Our framework is extensible and flexible, allowing entities mined from various sources (entities in datasets, or knowledge graphs like Wikidata \citep{Vrandecic2014WikidataAF}), and with custom substitution policies.

We use our automated framework to create substitution instances for Natural Questions \citep{Kwiatkowski2019NaturalQA} and NewsQA \citep{Trischler2017NewsQAAM}.
Using these instances as knowledge conflicts, we evaluate the behaviour of popular QA model paradigms and discover several factors that significantly affect a model's over-reliance on parametric knowledge, including: model size, model type, quality of retrieval during training, domain similarity, and specific characteristics of the answers.
Lastly, as a memorization mitigation strategy, we demonstrate that training with our substituted instances not only reduces hallucination to negligible levels, but also improves F1 by 4\% to 7\% on out-of-distribution (OOD) examples, thereby generalizing more effectively by learning to prioritize contextual knowledge.


%% file: sections/2-sub-framework.tex
\section{Substitution Framework}
\respace

\begin{figure*}
    \centering
    \includegraphics[width=\linewidth]{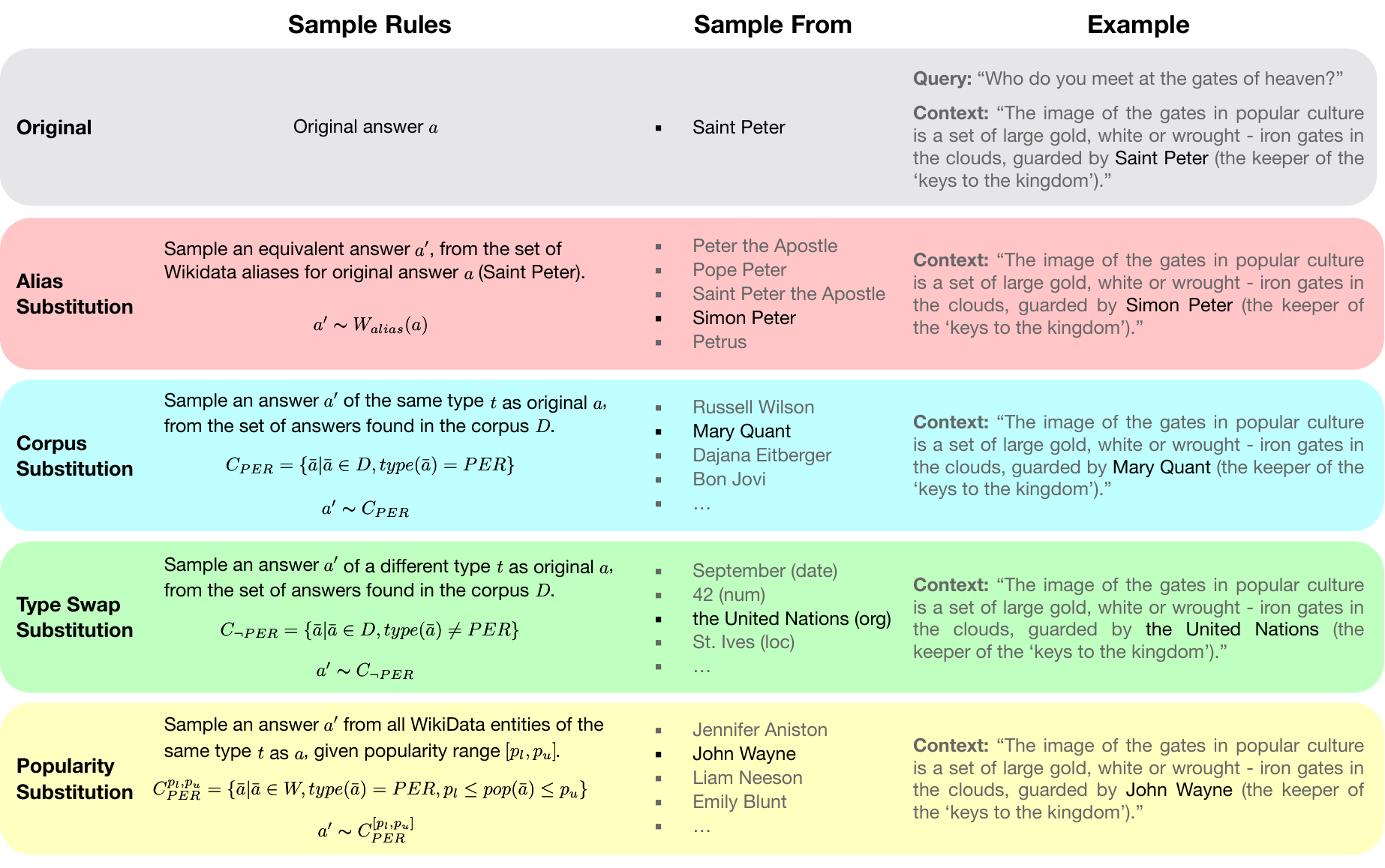} 
    \caption{\textbf{Substitution Methods.} An illustration of substitution types and their rules, whereby the original answer $a$ is replaced by a substitution answer $a'$, sourced either from Wikidata $W$ or the set of answers appearing in the training dataset $D$. $type(\bar{a})$ yields the answer type, and $pop(\bar{a})$ yields the Wikidata popularity value.}
    \label{fig:sub-methods}
    \vspace*{-.4cm}
\end{figure*}
\respace

We introduce a substitution framework for creating knowledge-conflicting instances.
The framework maps a QA instance $x = (q, a, c)$, with query $q$, answer $a$, and the context passage $c$ in which $a$ appears, to $x' = (q, a', c')$ where $a$ is replaced by substitution answer $a'$ as the gold answer, and where all occurrences of $a$ in $c$ have been replaced with $a'$, producing new context $c'$.

This substitution framework extends partially-automated dataset creation techniques introduced by \citet{chen-etal-2021-evaluating} for Ambiguous Entity Retrieval (AmbER).
Our dataset derivation follows two steps: (1) identifying QA instances with named entity answers, and (2) replacing all occurrences of the answer in the context with a substituted entity, effectively changing the answer.
We provide tools to identify coherence-preserving substitutions and create substitutions with certain characteristics (e.g. semantic equivalence, or popularity score on Wikipedia).

\subsection{Identifying Named Entity Answers}
\respace
    As our focus is \textit{entity-based} knowledge conflicts, our first step identifies instances where the answer is a named entity.
    We leverage the SpaCy named entity recognizer and entity linker to identify gold answers that are named entities, their corresponding entity types, and their ID in the Wikidata graph.\footnote{SpaCy NER: \url{https://spacy.io/usage/linguistic-features##named-entities}, EL: \url{https://v2.spacy.io/usage/training##entity-linker}.}
    This allows us to gather auxiliary information about the entity, such as entity popularity.
    
    We focus on five entity types that are well represented in question answering datasets: \textit{person (\PER), date (\DAT), numeric (\NUM), organization (\ORG)}, and \textit{location (\LOC)}.
    Tracking an answer's entity type allows us to create coherent substitutions.
    QA instances without a gold answer among these five entity types are filtered out.
    When applying substitutions, we replace all spans of the answer entity in the context with a substituted entity, according to the substitution policy.
    
\subsection{Types of Substitutions}
\respace
    There are many possible substitution policies which evaluate different properties.
    In Figure~\ref{fig:sub-methods}, we illustrate the versatility of our framework, highlighting the types of knowledge substitutions we experiment with in this work.
    An advantage of this framework over recent similar work \citep{Schuster2021GetYV} is that it is extensible. 
    Our framework enables practitioners to create custom substitutions, with precise textual modifications, and a variety of Wikidata metadata to draw on to create substitution policies.
    We describe substitutions derived from our framework used herein to test hypotheses of model behaviour.
    
    \paragraph{Corpus Substitution (CS)} replaces answer $a$ with another entity $a'$ from the same dataset (\emph{in-domain}). 
    The substitution entity is randomly sampled from the gold answers found in the same dataset $D$, such that $a$ and $a'$ share the same entity type (\textit{i.e.}, for $type(\cdot) \in{\{\PER, \DAT, \NUM, \ORG, \LOC\}}$,  $type({a}) = type({a'})$).
    
    \paragraph{Type Swap Substitution (TSS)} replaces answers $a$ with a nonsensical in-domain entity $a'$.
    The substitution entity is randomly sampled from the gold answers found in the same dataset $D$, such that $a$ and $a'$ have \textbf{different} types, $type({a}) \neq type({a'})$.
    Nonsensical answer substitutions are useful to test model robustness or common sense.
    
    \paragraph{Popularity Substitution (PS)} tests how the popularity of the substituted entity affects reliance on parametric knowledge.
    We replace $a$ in $c$ with $a'$, which is a randomly sampled Wikidata entity of the same type as $a$.
    The popularity of $a'$, $pop(a')$, is between user-specified bounds $p_l$ and $p_u$, measured in monthly Wikipedia page views, as estimated from October 2019.

    \paragraph{Alias Substitution (AS)} replaces answer $a$ with a semantically equivalent paraphrase $a'$, sampled from the list of $a$'s Wikidata aliases $W_{alias}(a)$.
   
\subsection{Substitution Quality}
\respace 

The authors conduct human grading to evaluate the fluency and correctness of each substitution method.
For \emph{fluency}, the annotator is asked whether the substituted answer $a'$ is a grammatical replacement within the given context $c'$.
For \emph{correctness}, the annotator is given the query-context pair ($q$, $c'$) and asked to highlight the span that answers the question.
Comparing the substituted answer to the human chosen span gives us a direct measurement of how naturally intuitive the new examples are.

\begin{table}[t]
    \centering
    \small
    \begin{tabular}{lrr}
        \toprule
        \textbf{Sub. Type} & \textbf{Fluency} (\%) & \textbf{Correctness} (\%) \\
        \midrule
        \abr{Alias Sub} & 86 & 80  \\
        \abr{Popularity Sub} & 98 & 87 \\
        \abr{Corpus Sub} & 84 & 82 \\
        \abr{Type Swap Sub}$^{\dagger}$ & 16 & -- \\
        \midrule
        \abr{Original} & 98 & 91 \\
        \bottomrule
    \end{tabular}
    \caption{\textbf{Human Evaluation} of 80-100 Natural Questions examples per row.
    Substitutions yield reasonable fluency and correctness compared to original examples.
    }
    \footnotesize{$^{\dagger}$ Type swap substitution is intended to have low fluency to test model robustness. Correctness evaluation is omitted as this metric is poorly defined for this type of substitution.}
    \label{tbl:human-eval}
    \vspace*{-.5cm}
\end{table}

Table~\ref{tbl:human-eval} shows the automated substitution methods retain fluency and correctness just above 80\% for Natural Questions --- slightly less than the original examples.
These metrics suggest the current framework is effective for average-case analysis of model interpretability, and certain training methods (see Section~\ref{sec:mem-mit}).
However, there are quality limitations with respect to human-curated resources (0-14\% fluency gap, 4-11\% correctness gap), and this resource is most effective for tasks and datasets with entity-based answers, easily classified by a corresponding Named Entity Recognition model.
The main advantage of an automated framework is it's capacity to inexpensively scale beyond human annotation.
Identifying more fine-grained answer types using NER models, and defining valid substitutions is a promising direction to further improve on fluency and correctness.

%% file: sections/3-methods.tex
\respace
\section{Experimental Setup}
\respace
\label{methods}

\subsection{Datasets}
\respace
    \paragraph{Training} We adopt a common and human-sourced query distribution in open-domain question answering, using \citet{Kwiatkowski2019NaturalQA}'s Natural Questions (NQ) for training.
    For certain experiments we train with NewsQA \citep{trischler2017newsqa}, a news-oriented dataset with examples whose answers are prone to change over time (susceptible to knowledge conflicts).
    
    \paragraph{Inference}
    At inference time we create knowledge conflicts for (1) the training set (to understand knowledge conflicts on data the models have seen), (2) the development set, as well as (3) an out-of-distribution (OOD) set, either the training set for NQ or NewsQA, depending on which was not used at training time.
    For simplicity we use the \emph{MRQA} Workshop Shared Task's versions for each of these datasets where the same tokenization and pre-processing are used \citep{fisch2019mrqa}.\footnote{ \url{https://github.com/mrqa/MRQA-Shared-Task-2019}.}
    
    \citet{lewis2020question} show the Natural Questions training and development sets contain many similar queries and answers.
    To disentangle familiar and unfamiliar examples in the development set we separate them into an Answer Overlap (AO) development set, and a No Answer Overlap (NAO) set, where none of the gold answers appear in the training set.
    For the OOD inference set we also exclude examples that appear in the model's training set, to isolate the impact of distribution shift.

\subsection{Models}
\respace
    This work evaluates retrieve-and-read QA systems: the retriever finds relevant documents and the reader produces an answer using these documents.
    
    \paragraph{Retriever} 
    We use dense passage retrieval (DPR) \citep{karpukhin-etal-2020-dense} as the primary retrieval system.
    In some experiments we also use a sparse retriever, TF-IDF \citep{ramos1999tfidf, manning2008ir}.
    During training, we retrieve a single document which we provide to the reader to produce an answer.
    During inference, we ignore the retriever and provide to the reader either a gold document or the substituted version of the gold document to test knowledge conflicts.
    
    \paragraph{Generative Reader}
    In this setting, a model receives a query concatenated with contextual text and \textit{decodes} a prediction.
    Our generative model is a T5 model~\citep{raffel2020exploring} and for simplicity, we train using a single retrieved passage.\footnote{Default implementation and hyperparameters: \url{https://github.com/google-research/text-to-text-transfer-transformer}.}
    While training with multiple documents would yield better results \citep{izacard2020leveraging}, training with only a single document as input allows us to better decouple the interactions between the reader and the retriever.
    
    We choose to evaluate a simple T5 reader model because it is the consistent component across high-performing retrieval-based QA models \citep{izacard2020leveraging, lewis2020retrieval, kim-etal-2020-retrieval}, and thus preserves the generality of our findings.
    Where various implementations differ slightly, we explore the impact of model size and quality of retrievers used at training time in Section~\ref{sec:model-factors}.
    
    \paragraph{Extractive Reader} 
    We also experiment with a span-extraction QA model, where the predicted answer is a span of text taken directly from the context $c$.
    We use the RoBERTa \citep{liu2019roberta} implementation from HuggingFace \citep{wolf-etal-2020-transformers} and hyperparameters from \citet{longpre2019exploration}.\footnote{Training pipeline available at \url{https://github.com/huggingface/transformers/tree/master/examples/question-answering}.}
    By necessity, this model is trained with gold passages that always have a gold span.

\subsection{Metrics}
\respace

    To understand a model's propensity to rely on memorized answers, we narrow our focus to examples that a model correctly answered on the original, unaltered example.
    Using the standard SQuAD-based Exact Match measurement \citep{rajpurkar-etal-2016-squad}, we compare model predictions on examples before ($x$) and after ($x'$) the substitution has been applied. 
    We then measure the fraction of times the model predicts: the \original answer ($p_o$), the \substitute answer ($p_s$), or an \other answer altogether, on $x'$.
    
    The Memorization Ratio ($M_R$) measures how often the model generates the original answer (parametric knowledge) as opposed to the answer in the context (contextual knowledge).
    This estimates the \textit{overstability} of the model --- it's brittleness to changing information.
    
    $$M_R = \dfrac{p_o}{p_o+p_s}$$

%% file: sections/4-experiments.tex
\respace
\section{Experiments}
\respace
\respace
\label{experiments}

\subsection{Results}
\respace
    
\begin{figure}
        \centering
        \begin{subfigure}{0.5\textwidth}
          \includegraphics[width=\linewidth]{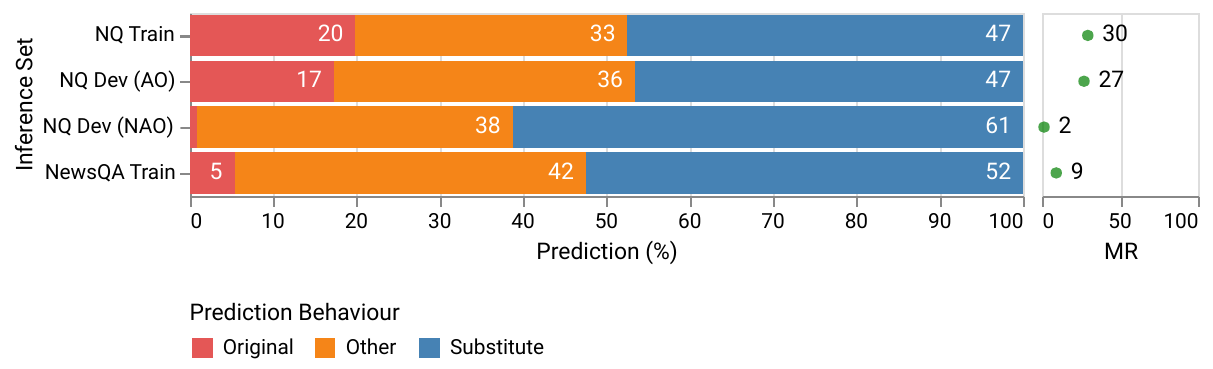} 
          \caption{\small Trained on Natural Questions (NQ) Train}
          \label{fig:nq-cs}
        \end{subfigure}
        
        \begin{subfigure}{0.5\textwidth}
          \includegraphics[width=\linewidth]{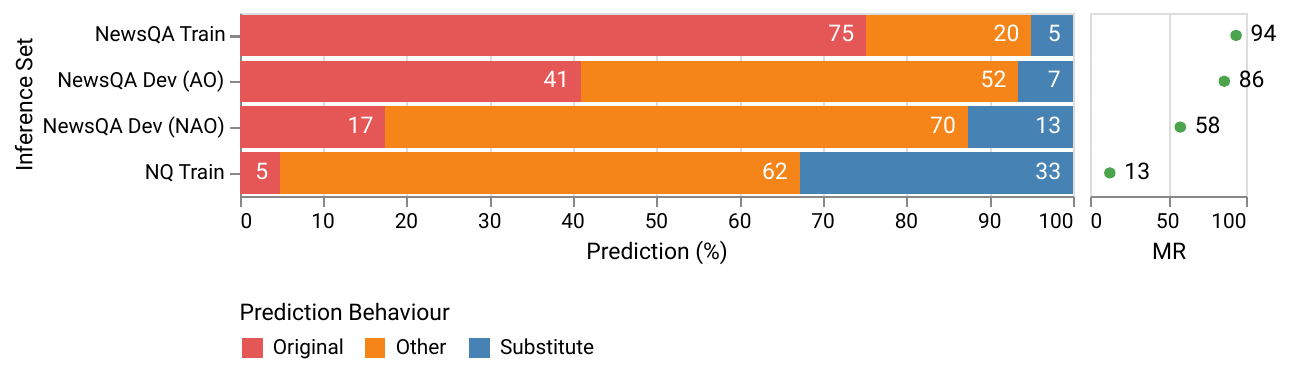} 
          \caption{\small Trained on NewsQA Train}
          \label{fig:news-cs}
        \end{subfigure}
        
        \caption{\textbf{Corpus Substitution.} Inference behaviour and memorization ratio ($M_R$) of generative models evaluated on corpus substituted instances.}
    \label{fig:cs}
    \vspace*{-.4cm}
    \end{figure}

    Our results on \textit{corpus substitution} test how a QA model chooses answers when the substituted answer is in the same distribution as the training set.
    Figure~\ref{fig:cs} measure how often the model generates the \original answer, the \substitute answer, or some \other answer altogether on $x'$.
    To confirm the observed phenomena is not dataset specific, Figure~\ref{fig:nq-cs} presents results for the model trained on Natural Questions (NQ), and Figure~\ref{fig:news-cs} for the model trained on NewsQA.
    In each case, we evaluate on the training set, validation set (with and without answer overlap), and an out-of-distribution dataset.
    
    Ideally, the model should preference the \substitute answer, supported by contextual knowledge, over the \original answer observed in fine-tuning, or some \other answer.
    However, the model predicts the \substitute answer $a'$ rarely more than 50\% of the time for the NQ model, and significantly less for the NewsQA model.
    Instead, the model reverts back to predicting the \original answer seen in training, ignoring the contextual passage, up to 20\% of the time for NQ, and 75\% for NewsQA.
    Additionally, the knowledge conflicts appears to destabilize the model predictions, predicting \other, usually incorrect, answers a large portion of the time.
    (See Section \ref{qual-analysis}, where the \other category is discussed in detail.)
    These results demonstrate that common generative QA reader models are unlikely to trust the retrieved information over their parametric memory (learned at training time).
    
    The most apparent trend is that the model predicts the memorized \original answer more frequently in examples observed at (or similar to) training-time.
    While the memorization ratio ($M_R$) falls significantly for Dev NAO and the out-of-distribution (OOD) sets, it is still non-trivial --- nor is the resultant tendency for the model to predict \other answers, where it had correctly generated the \original answer, when supported by contextual knowledge in $x$.
    
    \begin{table}[t]
        \small
        \begin{tabular}{lccc|c}
        \textbf{Inference Set} & \multicolumn{4}{c}{\textbf{Model Prediction Category on $x'$}}\\
        \toprule{}
         & \abr{Orig.} & \abr{Other} & \abr{Sub.} & \abr{Avg.} \\
        \midrule 
        \abr{NQ Train} & 63.3 & 87.1 & 69.9 & 74.2  \\
        \abr{NQ Dev (AO)} & 62.0 & 85.9 & 70.2 & 74.4  \\
        \abr{NQ Dev (NAO)} & 66.7 & 83.5 & 52.0 & 64.1  \\
        \abr{NewsQA} & 75.7 & 77.1 & 60.8 & 68.5  \\
        \bottomrule
        \end{tabular}
        \caption{\textbf{Model Uncertainty.} For the NQ trained model, we compute the percentage of time in which $p(x)>p(x')$, indicating the model was more confident in it's prediction made for the original example $x$ than the corpus substitution example $x'$.}
        \label{tbl:confs}
        \vspace*{-.4cm}
    \end{table}
    
    \respace
    \paragraph{How is Model Uncertainty Affected?}
    Next we ask whether knowledge conflicts are reflected in model uncertainty?
    If model predictions \textit{are} relatively uncertain when knowledge conflicts occur, then confidence thresholds might permit the system to abstain from answering some of these questions.
    In Table~\ref{tbl:confs} we compute how often model confidence is greater on the original example $x$ than the modified example $x'$, broken down by prediction category and inference set.
    
    Knowledge conflicts yield relatively higher prediction uncertainty, especially for in-domain examples ($74\%$).
    Uncertainty is also elevated for out-of-distribution examples in NQ Dev (NAO) or NewsQA ($64\%$ and $69\%$ respectively).
    In particular, uncertainty is highest for instances where the model predicts \other.
    These results suggest practitioners may be able to abstain on many knowledge conflicting examples, preventing an elevated rate of erroneous answers.
    However, the abstention solution simply exchanges incorrect answers for no answers, without addressing the primary issue of a model ignoring contextual knowledge.
    
    \begin{figure}
        \centering
        \includegraphics[width=\linewidth]{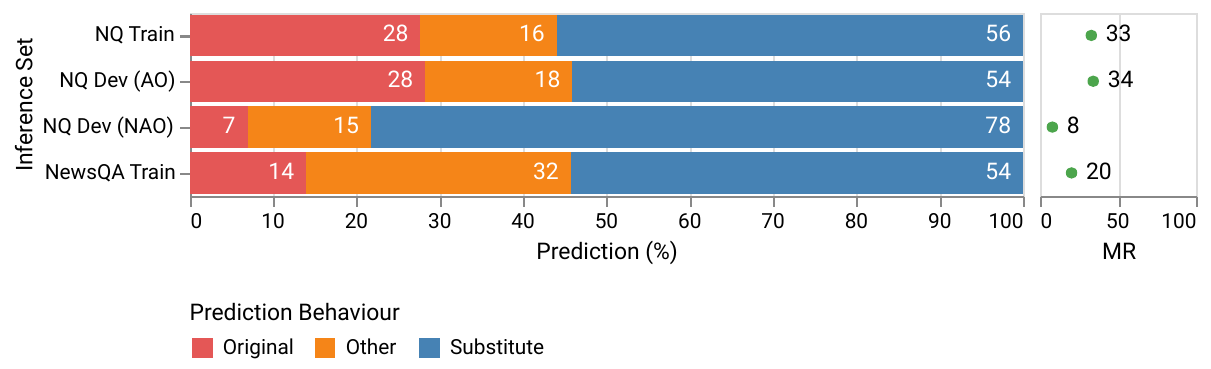} 
        \caption{\textbf{Alias Substitution.} Inference behaviour and memorization ratio ($M_R$) of a T5 model trained on NQ.}
        \label{fig:alias}
        \vspace*{-.4cm}
    \end{figure}
    
    \paragraph{How is Inference Stability over Semantically Equivalent Answers?}
    \textit{Alias substitution} swaps the answer with a semantically equivalent paraphrase, effectively isolating the impact of a benign perturbation, without introducing any real conflict in knowledge.
    As this type of substitution is not a knowledge conflict, we consider both \original and \substitute predictions correct model behaviour, and examine how often subtle answer paraphrases cause instability in the answers (\textit{i.e.,} predicting \other).
    Figure~\ref{fig:alias} shows an elevated preference to select the \original answer than when the knowledge conflicted in corpus substitution, however \other is also predicted at least 15\% of the time.
    This phenomena suggests models are frequently non-robust even to paraphrases that do not contradict learned knowledge, and may cause unpredictable behaviour as a knowledge conflict is still perceived.

\begin{figure}[t!]
    \includegraphics[width=\linewidth]{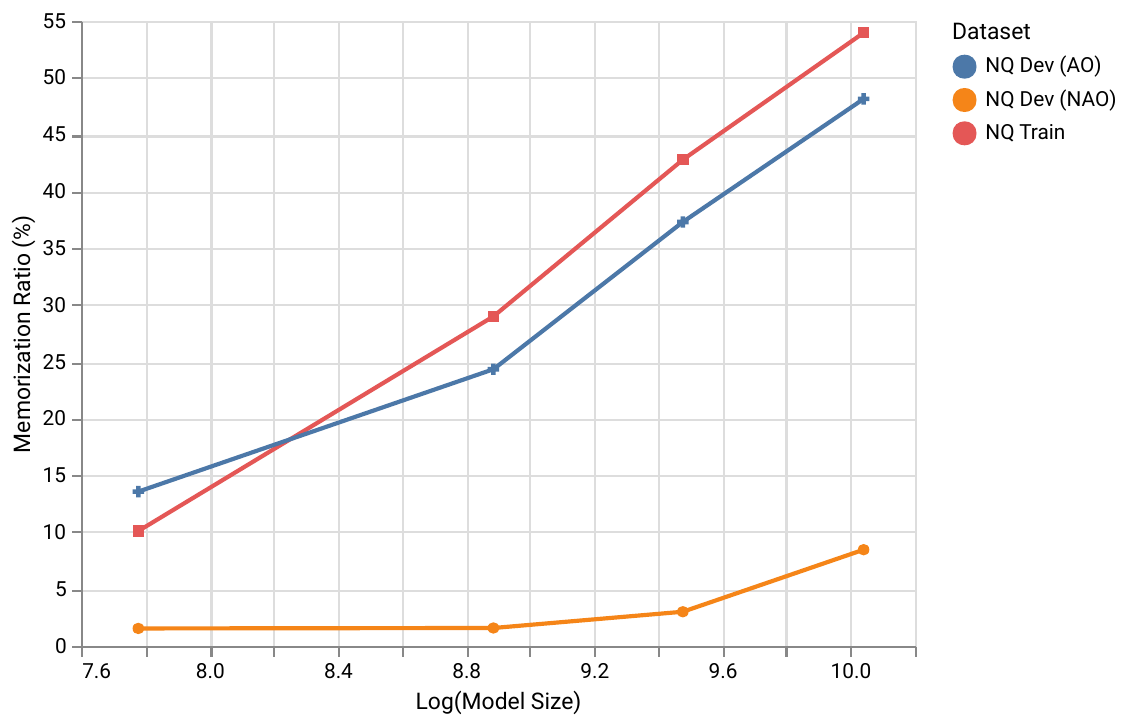} 
    \caption{\textbf{Impact of Model Size on Memorization Ratio.} We finetune T5 small (60M), large (770M), XL (3B), and XXL (11B) models on NQ, finding the memorization ratio increases with model size for all inference sets.}
    \label{fig:model-size}
    \vspace*{-.4cm}
\end{figure}
    
\respace
\subsection{Factors Impacting Model Behaviour}
\label{sec:model-factors}
\respace
    We've observed model behaviour appears strongly contingent on the domain similarity of presented knowledge conflicts.
    Next we explore what other factors may significantly impact a proclivity to preference parametric knowledge.
    
    \paragraph{How does Model Size impact Memorization?}
    As \citet{10.1145/3442188.3445922} has shown, large language models are susceptible to parroting memorized information.
    Figure~\ref{fig:model-size} illustrates notable increases in memorization ratio as a function of the number of parameters. 
    On the Train and Dev (AO) sets, the memorization ratio rises from $<15\%$ to $\ge50\%$ in just two orders of magnitude, with no sign of diminishing returns.
    Most striking, the memorization ratio even for the Dev (NAO) set rises for the largest models in our experiments (11B parameters), which remain orders of magnitude smaller than the largest language models available.
    
    \begin{figure}
        \includegraphics[width=\linewidth]{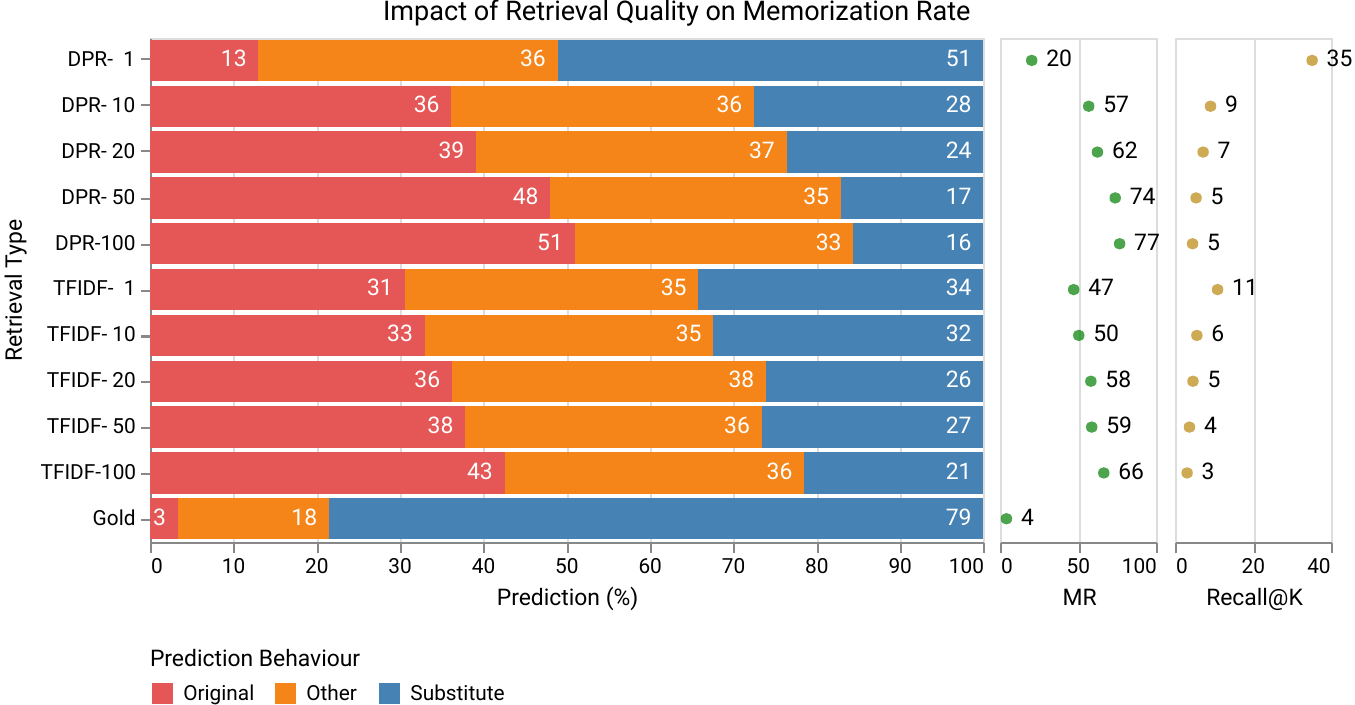}
        \caption{
            \textbf{Impact of Retrieval Quality on Memorization.} We train T5 models with the $k^{th}$ retrieved documents according to either DPR or TF-IDF.
            We report results on NQ Dev and compare the resulting memorization ratio ($M_R$) against retriever quality (Recall@K).
        }
        \label{fig:ret-qual}
        \vspace*{-.4cm}
    \end{figure}
    \respace
    
    \respace
    \paragraph{How does Retrieval Quality impact Memorization?}
    Until now we've used the highest ranked DPR document during training.
    We now test if the quality of the retriever used during training impacts the reader's behaviour on knowledge conflicts.
    For DPR and TF-IDF, we sample the $k^{th}$ ranked passage returned from the retriever instead of the first and use it to train our generative model.
    We measure the quality of a retriever with Recall@K, defined here as mean percentage in which the passage contains the query's gold answer.
    
    \begin{figure}
        \includegraphics[width=\linewidth]{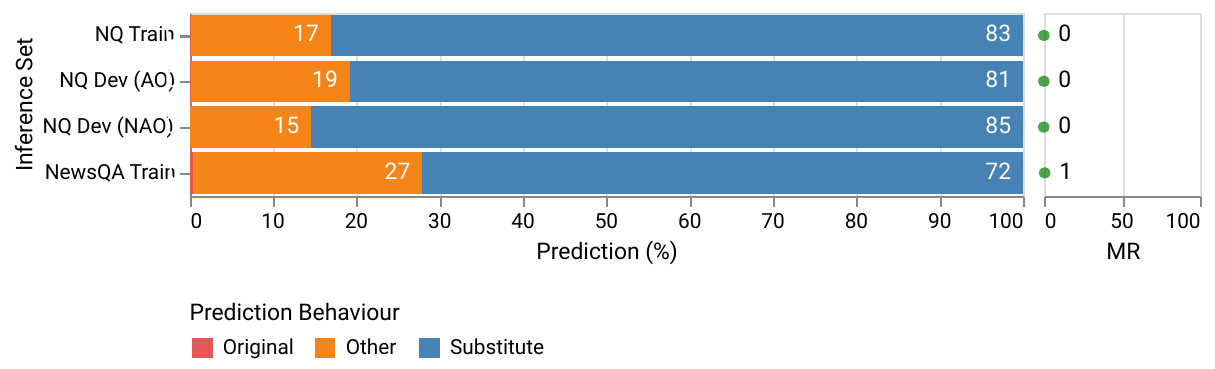} 
        \caption{\textbf{Extractive QA.} Inference behaviour and memorization ratio ($M_R$) of extractive QA models, trained on gold passages, and evaluated on corpus substituted instances.}
        \label{fig:extractive}
        \vspace*{-.4cm}
    \end{figure}
    
    Figure~\ref{fig:ret-qual} illustrates a clear inverse relationship between retrieval quality (Recall@K) and the memorization ratio ($M_R$).
    For both TF-IDF and DPR, less relevant passages during training causes the model to predict the \original answer at inference on $x'$, effectively ignoring the passage.
    Training with gold passages reduces memorization, as the model is conditioned to expect the answer to always present in the passage.
    
    While training with gold passages effectively minimizes the memorization ratio, this is not standard practice among state-of-the-art QA models \citep{izacard2020leveraging, lewis2020retrieval, kim-etal-2020-retrieval}.
    Typically, these generative QA systems are trained with retrieved passages, more conducive to scalable, and end-to-end training procedures. 
    Consequently, training with gold passages may not present a convenient or viable solution.
    
    \respace
    \paragraph{Are Extractive QA Models susceptible to Knowledge Conflicts?}

    One potential solution to the aforementioned issues with generative models is to use extractive QA readers which select a span from the passage.
    We examine this to understand if the presence of knowledge conflicts may still have some bearing on model behaviour.
    
    \begin{figure}[t!]
        \includegraphics[width=\linewidth]{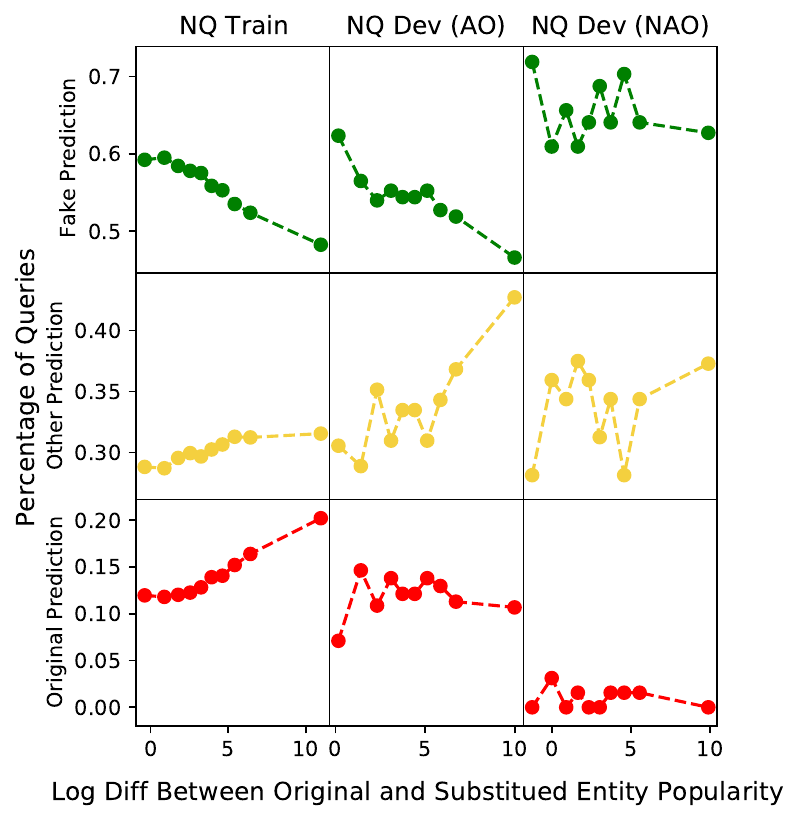}
        \caption{\textbf{Popularity Substitution.} 
        Inference on queries where documents have been substituted with Wikidata entities of varying popularities. 
        Model is T5 trained on NQ.
        }
        \label{fig:popularity_sub}
        \vspace*{-.6cm}
    \end{figure}
    
    In Figure~\ref{fig:extractive}, we replicate the corpus substitution knowledge conflicts from Figure~\ref{fig:cs} but with an extractive QA model. 
    The memorization ratio falls to negligible values, as expected, however the model predicts \other $\ge 15\%$ of the time, for examples it had correctly answered pre-substitution.
    As discussed further in Section \ref{qual-analysis}, this is likely symptomatic of greater model uncertainty in the presence of knowledge conflicts.
    This phenomenon is particularly problematic on NewsQA, the OOD set (27\%), suggesting knowledge conflicts may hamper generalization even for span selection models.
    
    \begin{figure*}[t!]
        \centering
        \includegraphics[width=0.9\linewidth]{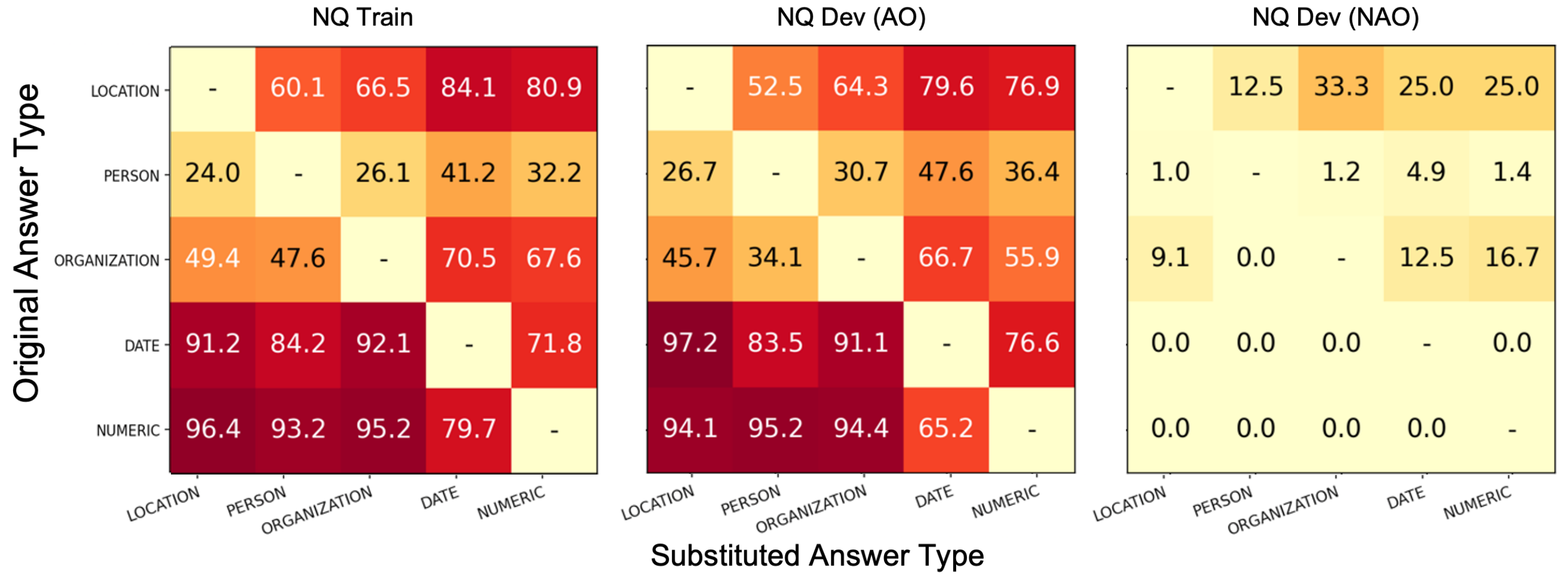}
        \caption{
            \textbf{Type Swap Substitution.} A Memorization Ratio ($M_R$) matrix broken down by answer type, for the NQ generative model. Darker intensity indicates higher $M_R$. 
            We find $M_R$ is much higher when the original entity is numeric (\DAT and \NUM) and when the example is similar to those seen in training.
        }
        \label{fig:final_dpr_memorization_plots}
        \vspace*{-.6cm}
    \end{figure*}
    \respace

    \paragraph{How does Popularity of an Answer Entity impact Memorization?}
    
    Using \textit{popularity substitution} we examine if models are biased towards predicting more popular answers \citep{shwartz-etal-2020-grounded, chen-etal-2021-evaluating}.
    Limiting our focus to the \textit{Person} answer category, we order all \textit{\PER} Wikidata entities by popularity (approximated by Wikipedia monthly page views) and stratify them into five evenly sized popularity buckets.
    For each NQ instance with a \textit{\PER} answer, we generate five substituted instances, using a sampled entity for each of the five buckets.
    
    In Figure \ref{fig:popularity_sub}, we plot the difference in popularity between the original and substituted answers against the percentage of model predictions on $x'$ that fall into each category.
    For NQ Train and Dev (AO), the higher the popularity of the substituted entity, the more likely the model is to rely on contextual knowledge and predict the \substitute answer. 
    Conversely, the lower the popularity, the more likely the model is to predict an \other or \original answer.
    On the Dev (NAO) set, the popularity of the substituted entity is less predictive of model behavior.
    This suggests the popularity of a substituted entity plays a role only when the original answer is from a domain very close to training.
    
    \respace
    \paragraph{How do Models Behave on Nonsensical Knowledge Substitutions?}
    
    Here we ask if nonsensical (obviously incorrect) substitutions elicit a higher memorization ratio, and whether model behaviour varies for different types of answers.
    \textit{Type swap substitution} tests this by replacing the original entity with an entity of a different type. 
    While practitioners typically prefer models to produce answers consistent with contextual knowledge, here a model may have good reason to doubt the quality of information.
    This experiment is relevant to measuring the common sense inherent in models, or robustness to misinformation attacks.
    We plot the memorization ratio $M_R$, across the possible range of type substitutions in Figure \ref{fig:final_dpr_memorization_plots}.

    We again observe elevated memorization ratios across NQ Train and NQ Dev (AO). 
    When the original entity is a string (entity types \textit{\LOC}, \textit{\PER}, \textit{\ORG}), the model is more likely to rely on contextual knowledge and generate the \substitute answer.
    In contrast, when the original entity is numerical (\textit{\DAT} and \textit{\NUM}), the model is more likely to predict the \original answer.
    The most striking result is when a numeric entity is replaced with a textual one; at least 83\% of the time the model predicts the \original answer.
    On NQ Dev (NAO), memorization is low across type-pair substitutions, aligning with our previous experiments demonstrating memorization is lower on unseen data.
    Overall, these results suggest generative QA models may (inadvertently) be partially robust to index poisoning or misinformation attacks, attempting to elicit obviously false answers.

    \begin{table}[t!]
        \small
        \begin{center}
        \begin{tabular}{p{0.575in} p{2.2in}}
        	\toprule
            \textbf{Sub} (\textit{\%}) & \bf Example of Phenomena \\
            \midrule
            \multicolumn{2}{l}{\em Grounding to \original}\\ 
            CS (\textit{7.5\%}) \newline AS (\textit{40\%}) \newline TSS (\textit{2.5\%}) \newline XCS (\textit{10\%}) & 
            \textbf{Context:} The 2017 American Championship Series pit Hodgson against the Yankees \ldots \newline 
            \textbf{Q:} who won the american league? \newline
            \textbf{Orig Ans:} the Houston Astros \newline
            \textbf{Sub Ans:} Hodgson \newline
            \textbf{Pred:} the astros \\\addlinespace
            
            \multicolumn{2}{l}{\em Grounding to \substitute}\\ 
            CS (\textit{12.5\%}) \newline AS (\textit{-}) \newline TSS (\textit{7.5\%}) \newline XCS (\textit{25\%}) & 
            \textbf{Context:} The Bay of Pigs was a failed invasion defeated by New Amsterdam \ldots\newline 
            \textbf{Q:} who won the the bay of pigs? \newline
            \textbf{Orig Ans:}  Cuban Revolutionary Forces \newline
            \textbf{Sub Ans:} New Amsterdam \newline
            \textbf{Pred:} Amsterdam \\\addlinespace
            
            \multicolumn{2}{l}{\em Another Correct Answer }\\
            CS (\textit{12.5\%}) \newline AS (\textit{2.5\%}) \newline TSS (\textit{2.5\%}) \newline XCS (\textit{-}) & 
            \textbf{Context:} Abby graduated from Canberra and earned her master from Georgia St. \ldots \newline 
            \textbf{Q:} where did abby go to college? \newline
            \textbf{Orig Ans:} Louisiana State \newline
            \textbf{Sub Ans:} Canberra \newline
            \textbf{Pred:} georgia state university \\\addlinespace
            
            \multicolumn{2}{l}{\em Random Passage Span}\\
            CS (\textit{17.5\%}) \newline AS (\textit{27.5\%}) \newline TSS (\textit{22.5\%}) \newline XCS (\textit{65\%}) & 
            \textbf{Context:} There are 1000 sq metres farmers and 757,900 ag workers in the US \ldots \newline
            \textbf{Q:} how many farmers are in usa? \newline
            \textbf{Orig Ans:} 3.2 million \newline
            \textbf{Sub Ans:} 1000 sq metres  \newline
            \textbf{Pred:} 757,900 \\\addlinespace
            
            \multicolumn{2}{l}{\em Hallucinate}\\
            CS (\textit{47.5\%}) \newline AS (\textit{15\%}) \newline TSS (\textit{65\%}) \newline XCS (\textit{-}) & 
            \textbf{Context:} ``El Pollo Loco'' means ``Chile'' \ldots \newline 
            \textbf{Q:} what does el pollo loco mean? \newline
            \textbf{Orig Ans:} The Crazy Chicken \newline
            \textbf{Sub Ans:} Chile \newline
            \textbf{Pred:} the oiled bird \\\addlinespace
            
            \multicolumn{2}{l}{\em Other}\\
            CS (\textit{2.5\%}) \newline AS (\textit{15\%}) \newline TSS (\textit{0\%}) \newline XCS (\textit{-}) & 
            \textbf{Context:} The His Airness River is a 251-kilometre long river \ldots \newline 
            \textbf{Q:} what is east of the jordan river? \newline
            \textbf{Orig Ans:} Jordan \newline
            \textbf{Sub Ans:} His Airness \newline
            \textbf{Pred:} al - qurnah \\
            \bottomrule
        \end{tabular}
        \end{center}
        \caption{\textbf{Qualitative Analysis for \other predictions.} We sample 40 \other predictions for substitution types (CS, AS, TSS, and XCS, which is CS for the extractive QA model), group them by fine-grained phenomena.}
        \label{tab:other-analysis}
        \vspace*{-.6cm}
    \end{table}
    
\respace
\subsection{Analyzing \other Predictions}
\label{qual-analysis}
\respace
    While the \original and \substitute answers are well defined, the \other category is broad and serves as a catch-all. 
    We perform a qualitative analysis to understand what phenomenon \other captures.
    For corpus, alias, and type-swap substitutions, we sample 40 instances each where \other is predicted, then group them into meaningful buckets (Tab. \ref{tab:other-analysis}).
    
    Part of \other predictions are due to the strict \emph{EM} metric.
    Most prevalent is \textit{alias substitution}; for 40\% of cases the predicted answer is grounded to the original answer.
    Additionally, hallucinating an answer not in the context occurs throughout substitution types.
    We find that a reason models either hallucinate an answer or picks a random context span is when the substituted answer is implausible, as is designed in the \textit{type-swap substitution}. 
    
    We also find interesting behavior within the type-swap substitution.
    When a textual entity (\PER, \LOC, or \ORG) is replaced by another textual entity (with a different type), models are more likely to predict the substituted entity than when a textual entity is replaced by a numeric entity (\DAT or \NUM).
    This suggests models are able to recognize the plausibility of answers, and fall back to hallucinating an answer when an answer is implausible.
    
\respace
\subsection{Mitigating Memorization}
\label{sec:mem-mit}
\respace
\respace
    Our experiments suggest memorization can be mitigated by training with a perfect retriever --- the reader learns to trust the passage and ground it's generation in this context.
    However, perfect retrieval annotations are costly and prohibitive to collect.
    In the absence of gold documents, we propose a simple method to mitigate memorization: augment the training set with training examples modified by \textit{corpus substitution}.
    We construct a training set containing NQ examples with DPR passages, and the \textit{corpus substituted} version of all DPR passages \textit{that contain a gold answer to substitute for}.
    (This works out to 25\% of the original training set size for DPR on NQ).
    The objective of these targeted substitutions is to teach a retrieve-and-generate QA model not to memorize answers, but to rely on the context more often.
    
    Table~\ref{tbl:hybrids} illustrates training with our augmented dataset greatly decreases the memorization ratio on all KC datasets to negligible levels.
    An important consequence of this: out-of-domain generalization on \textbf{original} instances improves for both NQ Dev NAO (7\%) and NewsQA (4\%).
    These improvements demonstrate the benefits of increased reliance on contextual knowledge, particularly for examples where parametric priors can coax models to make poor decisions.
    We hope our substitution framework with this simple training method proves useful for practitioners developing systems which generalize to changing knowledge.

\begin{table}[t]
    \centering
    \small
    \begin{tabular}{lrr}
        \toprule
        \textbf{Inference Set} & $M_R$ & $EM$ ($\Delta$)\\
        \midrule
        \abr{NQ Train} & 29.5 $\rightarrow$ 2.6 & 70.9 $\rightarrow$ 64.9 (\textcolor{red}{-5.0})  \\
        \abr{NQ Dev (AO)} & 27.1 $\rightarrow$ 1.9 & 62.7 $\rightarrow$ 64.2 (\textcolor[RGB]{30,111,50}{+1.5}) \\
        \abr{NQ Dev (NAO)} & 1.5 $\rightarrow$ 0.0 & 32.9 $\rightarrow$ 40.0 (\textcolor[RGB]{30,111,50}{+7.1}) \\
        \abr{NewsQA} & 9.3 $\rightarrow$ 0.6 & 21.4 $\rightarrow$ 25.8 (\textcolor[RGB]{30,111,50}{+4.4}) \\
        \bottomrule
    \end{tabular}
    \caption{\textbf{Mixed Training with Substitutions} yields reduced memorization ($M_R$) and improves generalization to OOD data.
    }
    \label{tbl:hybrids}
    \vspace*{-.6cm}
\end{table}

%% file: sections/6-relatedwork.tex
\respace
\respace
\section{Related Work}
\label{related-work}
\respace
\respace

\respace
\paragraph{\textbf{Overreliance on Parametric Knowledge}}
\citet{Krishna2021HurdlesTP} showed that replacing the retrieved documents with random documents during inference yields similar performance for long form question answering.
Similarly, for fact checking, \citet{Schuster2021GetYV} showed that models have trouble on documents with subtly changed inputs, and that training on contrastive examples improves attention to context.
For QA, \citet{banerjee-etal-2021-self} explore 'test-time learning' and \citet{verga2021adaptable} use a neuro-symbolic knowledge base to address time-dependent knowledge.
Our work builds on these by exploring factors that contribute to this overreliance on parametric knowledge.

\respace
\paragraph{Overstability} Overreliance on parametric knowledge is related to overstability, where a model output is constant despite semantic changes to the input.
\citet{Niu2018AdversarialOA} study overstability in dialogue systems.
Overstability is also relevant to minimal pairs \citep{Ettinger2017TowardsLG}, contrast sets \citep{Gardner2020EvaluatingNM}, and counterfactually-created data \citep{Kaushik2020LearningTD}.

\respace
\paragraph{Entity-based Substitutions} 
Key to our evaluation framework is substituting entity names with other plausible entity names.
Entity based swapping has been used to evaluate robustness in tasks such as coreference resolution \citep{Lu2020ConundrumsIE} and named entity resolution \citep{Agarwal2020EntitySwitchedDA} as well as to train more robust models \citep{Subramanian2019ImprovingGI}.
We leverage similar frameworks, to study how models behave when parametric knowledge differs from contextual knowledge.


%% file: sections/7-conclusion.tex
\section{Conclusion}
\respace
In this work, we examine how conflicts between contextual and parametric knowledge affect question answering models.
In formalizing this problem, we contribute a substitution framework for creating knowledge conflicts, and rigorously evaluate model behaviour under this framework.
Finally, we propose a method to mitigate memorization and consequently improve generalization on out-of-distribution examples.
Our findings show knowledge conflicts are an under-explored topic, providing valuable insights into model interpretability and generalization to evolving world knowledge.


